%% file: main.tex
\documentclass[10pt,twocolumn,letterpaper]{article}

\usepackage{soul}   

\newcommand{\sysname}{\textsc{RampNet}\xspace} 

\usepackage{iccv}              

\usepackage[english]{babel}
\usepackage[autostyle, english = american]{csquotes}
\usepackage[accsupp]{axessibility}
\MakeOuterQuote{"}


\definecolor{iccvblue}{rgb}{0.21,0.49,0.74}
\usepackage[pagebackref,breaklinks,colorlinks,allcolors=iccvblue]{hyperref}

\def\paperID{6} 
\def\confName{ICCV}
\def\confYear{2025}

\usepackage{graphicx}
\usepackage{cuted}
\usepackage{capt-of}
\usepackage{stfloats}

\title{\sysname: A Two-Stage Pipeline for Bootstrapping Curb Ramp Detection in Streetscape Images from Open Government Metadata}

\author{
    \setlength{\tabcolsep}{0.7em}
    \begin{tabular}{ccc}
        \textbf{John S. O'Meara}\textsuperscript{1,2} & \textbf{Jared Hwang}\textsuperscript{2} & \textbf{Zeyu Wang}\textsuperscript{2} \\
        \multicolumn{3}{c}{\textbf{Michael Saugstad}\textsuperscript{2} \qquad \textbf{Jon E. Froehlich}\textsuperscript{2}} \\
    \end{tabular}
    \\[2ex]
    \textsuperscript{1}Issaquah High School \qquad \textsuperscript{2}University of Washington \\
    \\\url{https://github.com/ProjectSidewalk/RampNet}
}

\begin{document}
\maketitle


\begin{strip}
  \centering
  \includegraphics[width=\textwidth]{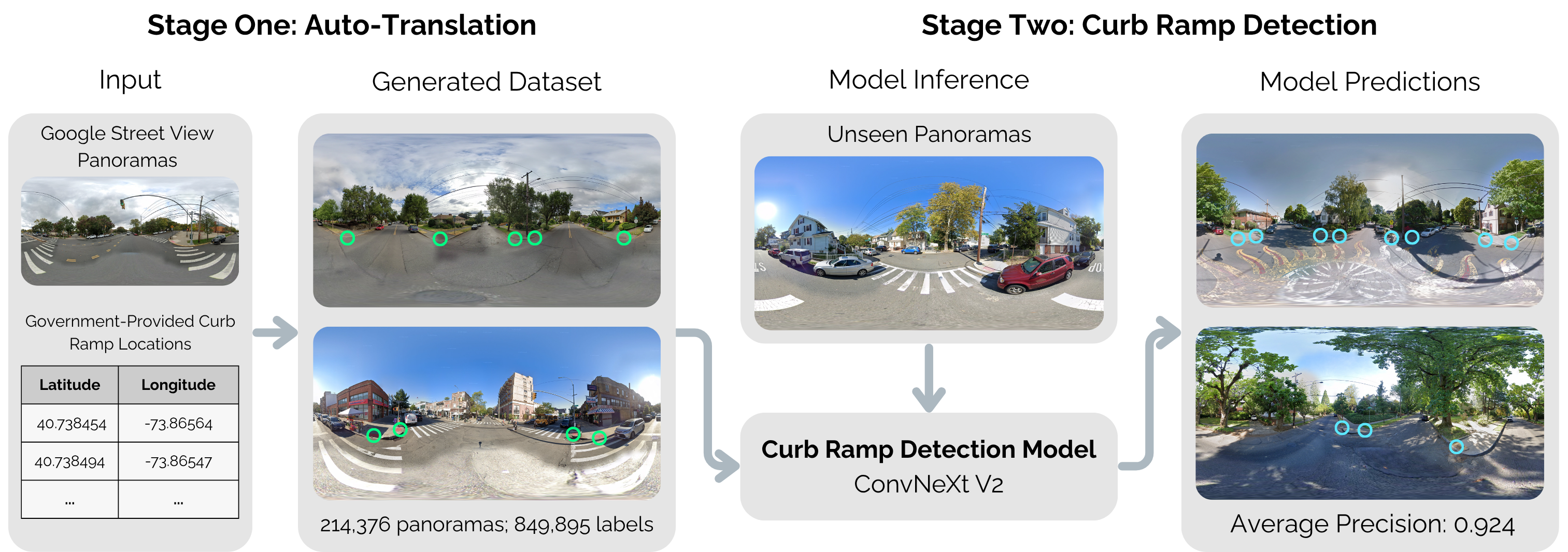}
  \captionof{figure}{We introduce \sysname, a custom two-stage pipeline for bootstrapping curb ramp detection models in streetscape images from open government data. In Stage 1, we auto-generate a labeled streetview curb ramp dataset by translating government-provided curb ramp location data (\ie, \textless{}lat, long\textgreater{} lists) into pixel labels on Google Street View panoramas. Stage 2 then uses this generated dataset to train a detection model that predicts curb ramp points in unseen panoramas.}
  \label{fig:teaser}
\end{strip}

\input{sec/0_abstract}    
\input{sec/1_intro}
\input{sec/2_related_work}
\input{sec/3_stage_one}
\input{sec/4_stage_two}
\input{sec/6_discussion}
\input{sec/7_conclusion}
\input{sec/8_acknowledgements}
{
    \small
    \bibliographystyle{ieeenat_fullname}
    \bibliography{main}
}

\end{document}

%% file: sec/0_abstract.tex
\begin{abstract}
Curb ramps are critical for urban accessibility, but robustly detecting them in images remains an open problem due to the lack of large-scale, high-quality datasets. While prior work has attempted to improve data availability with crowdsourced or manually labeled data, these efforts often fall short in either quality or scale. In this paper, we introduce and evaluate a two-stage pipeline to scale curb ramp detection datasets and improve model performance. In Stage 1, we generate a dataset of more than 210,000 annotated Google Street View (GSV) panoramas by auto-translating government-provided curb ramp location data to pixel coordinates in panoramic images. In Stage 2, we train a curb ramp detection model (modified ConvNeXt V2) from the generated dataset, achieving state-of-the-art performance. To evaluate both stages of our pipeline, we compare to manually labeled panoramas. Our generated dataset achieves 94.0\% precision and 92.5\% recall, and our detection model reaches 0.9236 AP---far exceeding prior work. Our work contributes the first large-scale, high-quality curb ramp detection dataset, benchmark, and model. 
\end{abstract}

%% file: sec/1_intro.tex

\section{Introduction}
\label{sec:intro}

Curb ramps are critical to urban accessibility~\cite{rosenbergOutdoorBuiltEnvironment2013, mrakRoleUrbanSocial2019}, impacting both independence~\cite{newtonIncreasingIndependenceOlder2010} and wellbeing~\cite{smithMobilityBarriersEnablers2021} for people with mobility disabilities. However, information regarding curb ramp location and condition is limited, making it difficult to plan and maintain accessible infrastructure~\cite{deitzSqueakyWheelsMissing2021, froehlichGrandChallengesAccessible2019}. To obtain this information, municipalities currently perform ``\textit{boots-on-the-ground}'' sidewalk inspections~\cite{stollofPedestrianMobilitySafety2008}, which can be prohibitively expensive and time-consuming. To address these challenges, computer vision-based techniques using streetscape images have been attempted~\cite{weldDeepLearningAutomatically2019, duanScalingCrowd+AISidewalk2022, haraTohmeDetectingCurb2014, adamsTrainingComputersSee2022}, but their results are limited by a lack of high-quality datasets~\cite{weldDeepLearningAutomatically2019}. While great strides have been made in the object detection space, curb ramp detection remains difficult due to data scarcity.

Existing curb ramp detection datasets are unsuitable for training deep learning models. \textit{Project Sidewalk}~\cite{sahaProjectSidewalkWebbased2019} is a crowdsourcing initiative that allows users to identify and assess curb ramps and other accessibility features through a Google Street View (GSV) web interface. While this approach is scalable, its design does not require users to comprehensively label each panorama (pano), rendering the data suboptimal for deep learning methods. Indeed, Weld \etal discuss this "under-labeling" as a limitation of using Project Sidewalk data to train object detection models~\cite{weldDeepLearningAutomatically2019}. Despite its limitations, the Project Sidewalk dataset~\cite{ProjectSidewalkSidewalkcvassets192025} is currently one of the only publicly available data sources for image-based curb ramp detection. One alternative is the \textit{Mapillary Vistas} dataset~\cite{neuholdMapillaryVistasDataset2017}, which has a class for "curb cuts"; however, we found that their categorization was overly broad and included driveways labeled as curb cuts.

In this paper, we introduce and evaluate a two-stage pipeline (\cref{fig:teaser}), called \textit{\sysname}, that leverages open government-provided curb ramp location datasets to auto-label curb ramps in streetscape images. While many local governments lack open data on curb ramp locations~\cite{deitzSqueakyWheelsMissing2021}, the datasets that are available contain only metadata (\eg, text lists of \textless{}lat, long\textgreater{}) without corresponding images, precluding their usage with computer vision techniques. Thus, in Stage 1 of our pipeline, we introduce an automated method to translate these city-collected location coordinates to image pixel coordinates in GSV panoramas, allowing us to create a dataset comprising over 840,000 auto-generated curb ramp labels. During the translation process, we use a \textit{ConvNeXt V2}~\cite{wooConvNeXtV2Codesigning2023} deep learning model to isolate the exact point of a curb ramp given a crop in the direction of the object's location (derived from the \textless{}lat, long\textgreater{}). In Stage 2, we use our generated dataset to train a separate deep learning model (again using ConvNeXt V2) that detects curb ramps in GSV panoramas, demonstrating real world applicability. By introducing a highly scalable, automatic, image-based curb ramp labeling pipeline, along with a new open dataset and initial benchmarks, our overarching goal is to help standardize and advance curb ramp detection research---similar to how \textit{WIDER FACE}~\cite{Yang_WiderFace_CVPR2016} and \textit{VGGFace2}~\cite{Cao_VGGFace2_FG2018} advanced face detection research.

We evaluate both pipeline stages by comparing to manually labeled panoramas (ground truth). We randomly selected and manually labeled 1,000 panoramas, yielding 3,919 manual curb ramp labels. In comparing our Stage 1 output to ground truth, we find a high level of agreement with the manual annotations, correctly identifying 92.5\% of curb ramps with 94.0\% precision. For Stage 2, we find that our model achieves state-of-the-art performance (0.924 AP) for curb ramp detection, greatly surpassing a previous method~\cite{weldDeepLearningAutomatically2019} that relied solely on crowdsourced data.

In sum, our research contributes: (1) a new technique for translating government-provided curb ramp location data (\textless{}lat, long\textgreater{} lists) to pixel locations in streetscape panoramas; (2) the first large-scale, high-quality curb ramp detection dataset; (3) a comprehensive benchmark for measuring curb ramp detection performance; and (4) a state-of-the-art curb ramp detection model. Our code and datasets are also open source\footnote{https://github.com/ProjectSidewalk/RampNet}, allowing others to build off our work and establishing key benchmarks for the community. Importantly, while Stage 1 relies on pre-existing government metadata, this is only for bootstrapping. The Stage 2 model benefits \textit{all} cities with GSV availability.


%% file: sec/2_related_work.tex
\section{Related Work}
\label{sec:related_work}
We describe work in curb ramp auditing and automated streetscape analysis.

\subsection{Curb Ramp Audit Methods}

In the US, sidewalks are required to have curb ramps (or "curb cuts") to support mobility for all, including people in wheelchairs, caregivers pushing strollers, or even travelers pulling luggage~\cite{AmericansDisabilitiesAct, deitzSqueakyWheelsMissing2021}. Traditionally, cities audit curb ramps through \textit{"boots-on-the-ground"} sidewalk inspections~\cite{stollofPedestrianMobilitySafety2008}. However, these manual audits are expensive and time-consuming~\cite{rundleUsingGoogleStreet2011}. Indeed, in a recent study of 178 US cities, Deitz \etal found that only 10\% published data on curb ramps~\cite{deitzSqueakyWheelsMissing2021}. Recent work has attempted to accelerate this auditing process with handheld LiDAR devices to reduce the amount of measurements performed manually~\cite{olsenPocketLidarCurb2024, aiImprovingPedestrianInfrastructure2019, turkanAutomatedLocalizationFunctional2022a,DeepWalkADASolutions}. However, these tools still require workers to conduct on-site data collection, limiting scalability. Moreover, this approach focuses primarily on assessing the quality of curb ramps at known locations, rather than detecting \textit{where} curb ramps are, as we do in this work. In general, in-person data collection still faces challenges with regards to cost and scalability despite attempts to improve efficiency. Some municipalities offer community-oriented tools~\cite{PedestrianRampComplaint} that allow residents to report accessibility issues, but these systems again rely on \textit{in situ} observation and active citizen participation, limiting scalability.


Prior work has attempted to leverage sensor data (\eg, accelerometer, gyroscope, and magnetometer data) to aid in urban accessibility assessment. \textit{Briometrix}~\cite{WheelchairPilotsBriometrix}, a company that specializes in sidewalk mobility audits, uses instrumented wheelchairs to collect sensing data. Similarly, \textit{SideSeeing}~\cite{damacenoSideSeeingMultimodalDataset2024}, a multimodal dataset for sidewalk assessment, uses video and sensing data captured with chest-mounted mobile devices. Unlike traditional in-person sidewalk audits, this approach directly involves pedestrians in the auditing process. Still, the need for physical, on-site data collection and inspection hinders scalability.

Virtual crowdsourcing initiatives are a promising alternative~\cite{haraCombiningCrowdsourcingGoogle2013}. For example, Project Sidewalk~\cite{sahaProjectSidewalkWebbased2019} allows users to identify and assess accessibility features like curb ramps through a GSV web interface. Compared to field audits, this technique scales rapidly and at a lower cost. Importantly, users need not be physically present in the city to contribute. Previous work has demonstrated a high level of agreement between these crowdsourced labels and government field data~\cite{askariValidatingPedestrianInfrastructure2025}. While Project Sidewalk's crowdsourcing approach scales, it is still limited due to its requirement for human labor. In addition, Project Sidewalk does not require users to comprehensively label panoramas, complicating efforts to use its data in deep learning applications~\cite{weldDeepLearningAutomatically2019}.

Others have explored computer vision techniques to detect pedestrian features in aerial imagery~\cite{hosseiniMappingWalkScalable2023}, including for sidewalks~\cite{hosseiniGlobalScaleCrowd+AITechniques2022, ningSidewalkExtractionUsing2022}, crosswalks~\cite{ahmetovicMindYourCrossings2017}, and roads~\cite{shamsolmoaliRoadSegmentationRemote2021}. However, detecting curb ramps~\cite{UsingGeoAIInventory} remains challenging due to occlusions from shadows and trees, their relatively small size, and their tendency to visually blend in with sidewalks. It is also difficult to reliably assess curb ramp quality factors (\eg, steepness, presence of tactile warnings) due to the low resolution and inherent limitations of aerial imagery. While our work focuses on detection rather than quality assessment, our usage of high-resolution streetscape imagery instead of aerial imagery enables future work for this task.

\subsection{Automated Streetscape Analysis}

With the rise of computer vision and deep learning technologies, streetscape imagery has increasingly become an important tool for large-scale urban analysis~\cite{biljeckiStreetViewImagery2021, hou2024global}. GSV, the largest repository of such imagery, primarily collects panoramas along public roadways, offering a plethora of data about the built environment. This data has been used in multiple domains, including accessibility assessment~\cite{weldDeepLearningAutomatically2019, duanScalingCrowd+AISidewalk2022, haraTohmeDetectingCurb2014, adamsTrainingComputersSee2022}, demographic studies~\cite{UsingDeepLearning}, neighborhood quality assessment~\cite{wang2024assessing}, and real estate valuation~\cite{lawTakeLookUsing2019}. When used in conjunction with computer vision techniques, streetscape imagery provides a low-cost and scalable method for understanding urban environments.

Prior work has attempted to leverage streetscape imagery and computer vision models to detect curb ramps and other accessibility features in images. For example, Hara \etal developed \textit{Tohme}~\cite{haraTohmeDetectingCurb2014}, a system that combines crowdsourcing and computer vision to semi-automatically detect curb ramps in GSV imagery. While far from fully automated curb ramp detection, their system results in a 13\% reduction in time cost compared to manual labeling alone. Building on this work, Weld \etal introduced a customized \textit{ResNet} model trained on crowdsourced data to automatically detect accessibility features (\eg, curb ramps, missing curb ramps, obstructions, surface problems) in GSV panoramas~\cite{weldDeepLearningAutomatically2019}.


Despite these efforts, no existing curb ramp detection model comes close to human-level labeling performance. Tohme achieves a recall of 67\% and a precision of 26\% when benchmarked against manual labels~\cite{haraTohmeDetectingCurb2014}. Weld \etal improve upon this with a recall of 78.7\% and a precision of 33.7\%~\cite{weldDeepLearningAutomatically2019}, but these results are still infeasible for fully automatic urban accessibility assessment. While a variety of factors play into this poor performance, two key factors are: (1) the lack of large-scale, high-quality and open curb ramp image datasets; (2) the lack of standardized benchmarks. Our research attempts to address both, generating a dataset that is both larger and cleaner than previous efforts, and enabling a state-of-the-art curb ramp detection model that greatly outperforms prior work.

%% file: sec/3_stage_one.tex
\section{Stage 1: Dataset Generation}
\label{sec:stage_one}

We introduce and evaluate a two-stage pipeline (\cref{fig:teaser}), called \textit{\sysname}, for bootstrapping curb ramp detection in streetscape images from open government metadata. We first describe Stage 1, which auto-generates a labeled image-based curb ramp dataset by translating government-provided curb ramp location data to image pixel coordinates in GSV panoramas (\cref{fig:auto_translation}). Rather than traditional object detection datasets, which contain bounding boxes, we use single points to represent curb ramps, mirroring prior work in curb ramp detection~\cite{weldDeepLearningAutomatically2019, sahaProjectSidewalkWebbased2019}. At a high level, our dataset generation process involves identifying relevant panoramas, extracting directional image crops, localizing curb ramps within these crops, and aggregating the results back onto the full panorama.


\subsection{Dataset Selection}
We describe our dataset selection process both for the open government data as well as corresponding GSV panoramas.

\begin{figure}
  \centering
  \includegraphics[width=3.25in]{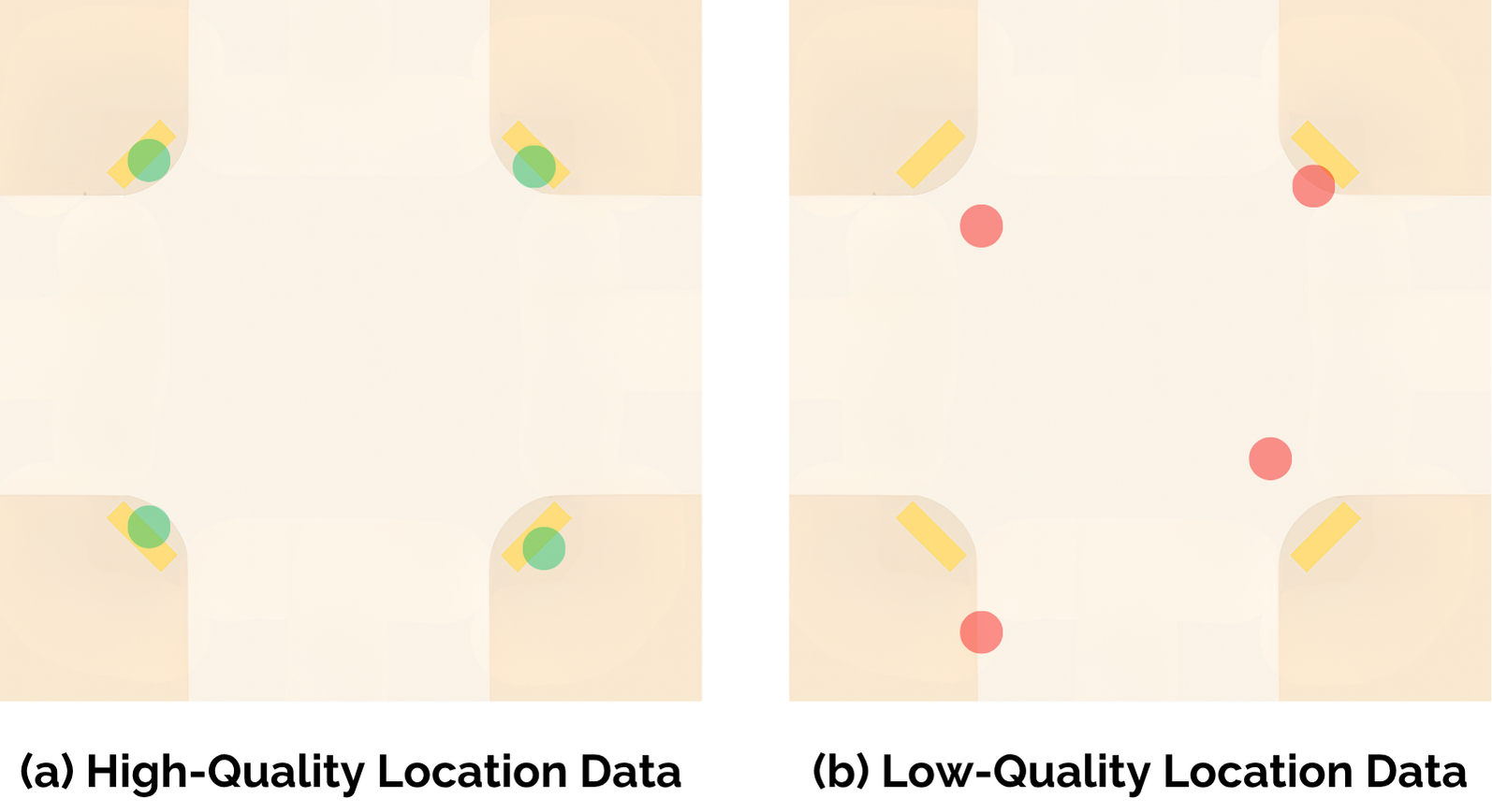}
  \caption{A visual comparison of location data quality. \textbf{(a)} High-quality coordinates precisely align with curb ramps' physical location. \textbf{(b)} In contrast, low-quality coordinates are often misplaced, rendering them unsuitable for our method.}
  \label{fig:location_quality}
\end{figure}

\textbf{Selecting government datasets.} Stage 1 is dependent on high quality curb ramp location data published by local governments. However, as noted previously, this data is rare: of the 178 US cities studied in \cite{deitzSqueakyWheelsMissing2021}, 90\% published open street data but only 34\% had sidewalk data and far fewer (10\%) included curb ramps. Our goal is to leverage those cities that \textit{do} publish curb ramp locations, translate those locations to pixels in GSV panos, and use this new dataset to train computer vision models.

Even when government curb ramp data is available, we observed location imprecision, which impacts our pipeline (see \cref{fig:location_quality}). To evaluate location quality, we manually examined curb ramp data from eight different local government datasets by overlaying curb ramp locations on top of an aerial imagery base map. Of the eight cities, one city (Seattle) had \textit{poor} location precision, four were \textit{OK}, and three were \textit{good}---see~\cref{tab:curb_ramp_loc_data_eval}. For our purposes, we use all data from the \textit{good} category: New York City, NY~\cite{PedestrianRampLocations}; Portland, OR~\cite{CurbRamps}; and Bend, OR~\cite{CurbRampsa}---all which offer precise and diverse curb ramp styles. 

\begin{table}[t]
  \centering
  {
    \begin{tabular}{@{}lll@{}}
      \toprule
      \small{City} & \small{\# of Curb Ramps} & \small{Location Precision} \\
      \midrule
      \small{Austin, TX} & \small{48,995} & \small{OK} \\
      \small{\textbf{Bend, OR}} & \small{\textbf{13,611}} & \small{\textbf{Good}} \\
      \small{Los Angeles, CA} & \small{91,759} & \small{OK} \\
      \small{Nashville, TN} & \small{18,285} & \small{OK} \\
      \small{\textbf{New York City, NY}} & \small{\textbf{217,680}} & \small{\textbf{Good}} \\
      \small{\textbf{Portland, OR}} & \small{\textbf{45,324}} & \small{\textbf{Good}} \\
      \small{Seattle, WA} & \small{45,653} & \small{Poor} \\
      \small{Washington D.C.} & \small{34,859} & \small{OK} \\
      \bottomrule
    \end{tabular}

    \vspace{2pt}
    
    {\small }
  }
  \caption{Initial sample of government-provided curb ramp datasets in the US and their location precision compared to aerial imagery.}
  \label{tab:curb_ramp_loc_data_eval}
\end{table}

While formats vary, each government-provided dataset includes at least a unique identifier and a \textless{}lat, long\textgreater{} tuple (\textit{e.g.,} see \cref{fig:teaser,fig:auto_translation}). For Stage 1, we need to convert these location tuples to corresponding pixel locations in a GSV panorama. For this, we need to determine which GSV panoramas best correspond to the curb ramp location data.

\textbf{Selecting GSV panoramas.} To identify and select corresponding GSV panoramas, we define a 10-meter radius around each government-defined curb ramp location and download all available panoramas within this boundary as an equirectangular image at $4096 \times 2048$ pixel resolution---a resolution we selected because both prior work~\cite{haraTohmeDetectingCurb2014} and our own testing conclude that further increasing the resolution results in minimal performance gain while significantly increasing computational complexity.

\textbf{Selecting label candidates.} For each panorama within the 10-meter radius, we then must determine which curb ramps to label within it. Using the government data, we select all curb ramps within a 35-meter radius of the panorama's location. This radius is intentionally larger than the 10-meter radius used for panorama selection, as it ensures that all curb ramps reasonably visible within the panorama are captured and labeled. Therefore, while a pano is selected based on a single nearby curb ramp that is within 10 meters, its final set of labels includes \textit{all} curb ramps within a larger 35-meter radius. We also require the curb ramp installation date (defined in the government data) to be earlier than the panorama capture date. For each panorama, all curb ramp locations that meet these constraints will be marked as label candidates and saved for the auto-translation step, where we convert their locations to image pixel coordinates.

To reflect real-world scenarios, we also infuse our dataset with 20\% null images (\ie, a panorama that contains zero curb ramps). To download a null panorama, we first randomly select one of the three cities: New York City, Portland, or Bend. We then choose a random point on that city's street network and locate nearby panoramas. If none of the nearby panos are positioned at least 60 meters away from any known curb ramp location, we repeat the process, continuing until we find a panorama that is at least 60 meters away from all curb ramps.

In total, the above process yields 219,170 panoramas and 959,442 label candidates. Of these panoramas, 54.9\% are from New York City, 8.5\% are from Bend, and 36.6\% are from Portland (left columns in  \cref{tab:datasets_used_in_stage1}). The discrepancy between cities can be attributed to differences in city size, GSV availability, and curb ramp quantity.

\subsection{Auto-Translating Geo to Image Coordinates}

\begin{figure*}[t]
  \centering
  \includegraphics[width=6.85in]{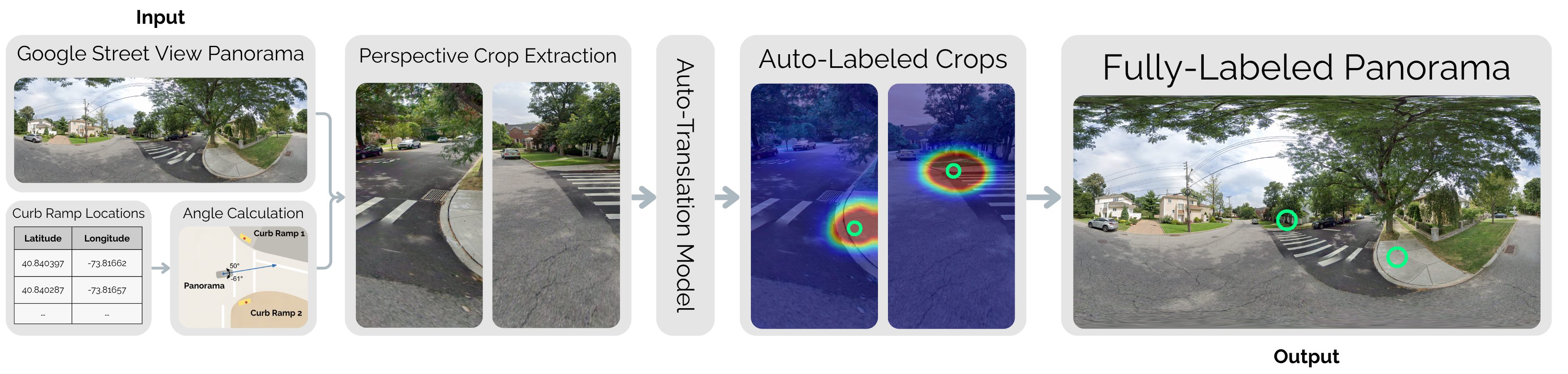}
  \caption{The auto-translation method used to generate our dataset. For each label candidate (a government-provided curb ramp location), we compute its angle with respect to the panorama location. With this angle, we extract a directional perspective crop from the full panorama. We then pass this crop to our auto-translation model, which outputs a heatmap whose peaks represent possible curb ramp points. These points are projected back onto the original equirectangular image, yielding a fully-labeled panorama.}
  \label{fig:auto_translation}
\end{figure*}

With the government-based curb ramp location data and corresponding GSV panos downloaded, we now need to translate the curb ramp \textless{}lat, long\textgreater{} coordinates to image pixel coordinates on the GSV panos. We do this in two parts: first, for each potential curb ramp in a pano, we extract a directional crop around the curb ramp in the pano and then, second, we use a trained ConvNeXtV2~\cite{wooConvNeXtV2Codesigning2023} model to isolate curb ramp points within these crops (\cref{fig:auto_translation}).

\textbf{Auto-cropping along curb ramp heading.} To generate the crops around each curb ramp, we calculate the angle between the panorama capture location and the curb ramp location. Using this angle, we generate a square perspective crop in the direction of the curb ramp ($1024 \times 1024$ pixels), akin to what a user would observe in GSV if they were looking toward it. The crop has a 90-degree field of view and a 30-degree downward pitch to better capture the ground level. To focus our analysis on relevant context, we retain only the middle third of the crop, producing a final image that is $341 \times 1024$ pixels. 

\textbf{Identifying curb ramp pixels.} To isolate the exact pixel coordinates of curb ramps within each crop, we use a modified ConvNeXt V2~\cite{wooConvNeXtV2Codesigning2023} model (specifically, the \textit{base} variant) that outputs a heatmap of probable curb ramp points. Our model is initially pretrained on crops extracted from 20,698 annotated Project Sidewalk~\cite{sahaProjectSidewalkWebbased2019} panoramas spanning 12 US cities, including Chicago, Seattle, Pittsburgh, and St. Louis. However, using Project Sidewalk data alone results in poor performance due to its many missing labels\footnote{As noted, users in Project Sidewalk are asked to label sidewalk features and obstacles in GSV imagery, including curb ramps; however, they may do so from multiple panoramas rather than comprehensively labeling a single panorama---which makes the data suboptimal for model training}. Indeed, a crop model that is solely trained on Project Sidewalk data alone achieves a recall of 76.7\%, and a precision of 77.2\%. To rectify this issue, we add a second round of training on crops extracted from a smaller, manually-labeled dataset of 312 panoramas from Bend, NYC, and Portland. Here, every curb ramp occurrence in each pano was labeled using Label Studio~\cite{LabelStudio}. By combining both data sources, our final crop model achieves a recall of 89.0\%, and a precision of 87.0\%. This crop-level model is not to be confused with our Stage 2 curb ramp detection model, which identifies curb ramps in whole panoramas, rather than just crops.

After the auto-translation process, our dataset contains 214,376 fully labeled panoramas and 849,895 curb ramp labels, larger than any previous efforts (see   \cref{tab:stage1_performance_comparison_with_ps}).

\begin{table}[t]
  \centering
  {
    \footnotesize 
    \begin{tabular}{@{}lllll@{}}
      \toprule
      \multicolumn{1}{c}{City} & \multicolumn{2}{c}{Initial} & \multicolumn{2}{c}{After Auto-Trans.} \\
      \cmidrule(lr){2-3} \cmidrule(lr){4-5}
      & Curb Ramp & Panoramas & Curb Ramp & Panoramas \\
      \midrule
      Bend & 20,451 & 18,545 & 19,082 & 18,205 \\
      New York & 655,084 & 120,430 & 566,832 & 117,323 \\
      Portland & 283,907 & 80,195 & 263,981 & 78,848 \\
      \midrule 
      \textbf{Total} & \textbf{959,442} & \textbf{219,170} & \textbf{849,895} & \textbf{214,376} \\ 
      \bottomrule
    \end{tabular}

    \vspace{2pt}
    
  }
  \caption{Table showing the number of initial curb ramp candidates and panoramas, and the final counts of successfully generated labels and panoramas after auto-translation in Stage 1.}
  \label{tab:datasets_used_in_stage1}
\end{table}

\subsection{Study Method}
We now describe our evaluation approach. Though our immediate focus here is on evaluating Stage 1 performance, we describe methodological details shared between Stage 1 and 2. We first divide our auto-labeled Stage 1 dataset of 214,376 panoramas into 70\% training, 20\% validation, and 10\% test splits. Because an individual curb ramp may be present in more than one panorama at differing angles or viewpoints, we must be wary of possible data leakage. To avoid this, we use a semi-random splitting strategy that groups panoramas located within 60 meters of each other into the same split. Our final experimental dataset contains 150,063 panoramas in the training split, 42,875 in validation, and 21,438 in test. While we do not use the validation split in this study, we provide it for other researchers for tasks such as hyperparameter tuning or model selection. We now describe our ground truth approach and correctness metrics; both are used to evaluate Stage 1 and 2.

\textbf{Ground truth.} To create a ground truth, we manually labeled curb ramps across 1,000 panoramas randomly sampled from the test split (independent of the 312 aforementioned manually labeled panoramas). We used Label Studio~\cite{LabelStudio} for this process. In total, we labeled the center points of 3,919 curb ramps (3.9 ramps/pano).

\textbf{Correctness metrics.} To evaluate the accuracy of predicted curb ramp labels against our manually-created ground truth, we use a proximity-based comparison method. For each predicted label, we check if there are any ground-truth points within an 88-pixel radius in the $4096 \times 2048$ pixel panorama. If not, we mark the label as a false positive. Conversely, if it does match one or more ground-truth points, those within the radius will be counted as a true positives. If more than one predicted label matches the same ground-truth point, we pick the one with highest confidence and ignore the others entirely in our calculation. This is to ensure that no ground-truth points are counted more than once. If any of the ground-truth points are not detected, we count these as false negatives.

\subsection{Results}
In comparing the 3,919 ground truth labels to Stage 1 output across 1,000 panoramas, our findings demonstrate a high level of agreement, achieving 92.5\% recall and 94.0\% precision for an F-score of 0.932. \cref{tab:stage1_performance_comparison_with_ps} compares our dataset to Project Sidewalk's \cite{sahaProjectSidewalkWebbased2019}, demonstrating significant improvements in both scale and comprehensiveness. 

\begin{figure*}[t]
  \centering
  \includegraphics[width=6.85in]{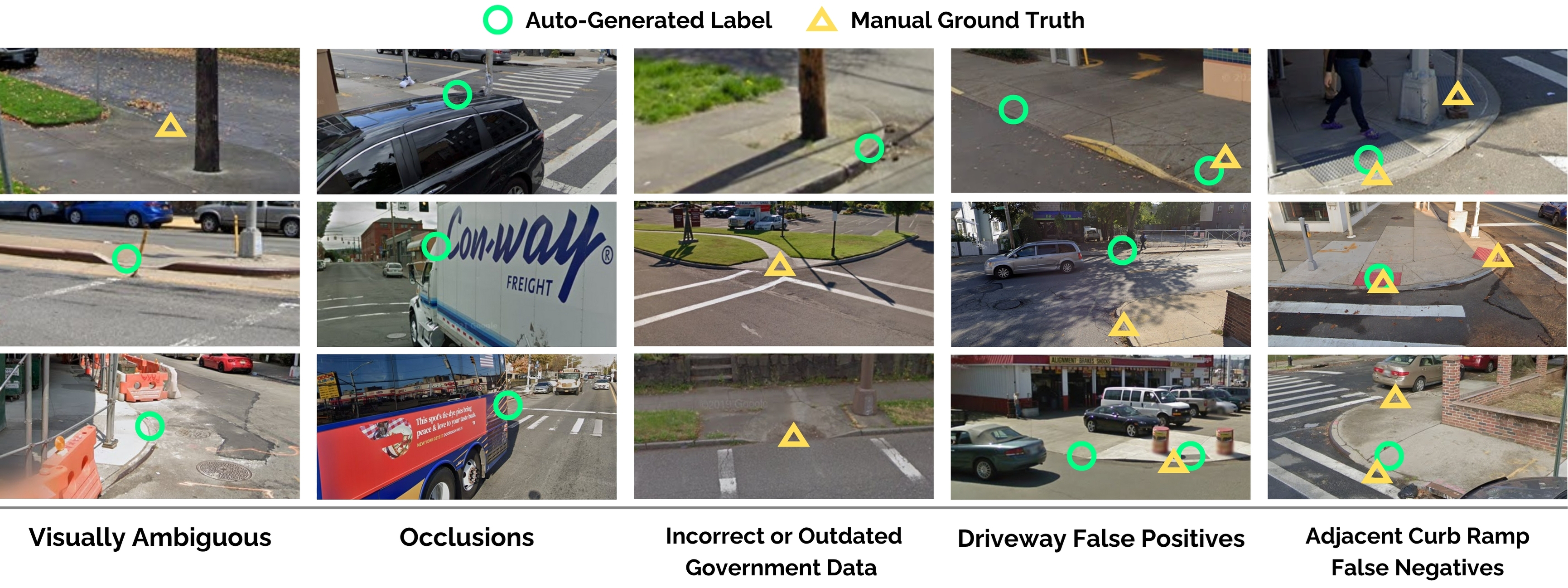}
  \caption{We performed a qualitative analysis of 100 randomly sampled Stage 1 errors and inductively categorized them into five groups: visually ambiguous (43\% of errors), occlusion (27\%), disagreements with government data (17\%), driveways mistaken for curb ramps (8\%), and unlabeled adjacent curb ramps (5\%).}
  \label{fig:qanalysis}
\end{figure*}

To advance understanding of performance, we randomly sampled and analyzed 100 errors and inductively coded them thematically. The most common error was disagreements on visually ambiguous curb ramps (43\%). These cases involved distant or low-resolution objects where a definitive identification is inherently subjective. Other common errors were the following: occlusions, where a car, bus, or other object obscures the curb ramp itself but our model still makes a speculative prediction (27\%); disagreements in government data either due to underlying errors in the open datasets (\textit{e.g.,} a missing curb ramp labeled as a curb ramp) or temporal differences between capture dates in the government metadata \textit{vs.} the GSV pano (17\%); driveways mistaken for curb ramps (\eg, an adjacent driveway leaks into a crop) (8\%); and unlabeled adjacent curb ramps (5\%). 

\begin{table}[t]
  \centering
  {
    \begin{tabular}{@{}lll@{}}
      \toprule
      & \small{Project Sidewalk} & \small{\sysname Stage 1} \\
      \midrule
      \small{Comprehensiveness} &  \small{Partially‑Labeled} & \small{Fully‑Labeled} \\
      \small{Data Source}            & \small{Crowdsourced} & \small{Auto‑Generated} \\
      \small{\# of Ramp Labels} & \small{427,435} & \small{849,895} \\
      \small{\# of Panoramas} & \small{149,574} & \small{214,376} \\
      \small{Labels / Panorama$^{*}$} & \small{2.86} & \small{4.96} \\
      \bottomrule
    \end{tabular}

    \vspace{2pt}
    
    {\small $^{*}$Excludes panoramas with zero labels.}
  }
  \caption{Comparison of Project Sidewalk and \sysname datasets.}
  \label{tab:stage1_performance_comparison_with_ps}
\end{table}

%% file: sec/4_stage_two.tex
\section{Stage Two: Curb Ramp Detection}
\label{sec:stage_two}

\begin{figure*}
  \centering
  \includegraphics[width=6.85in]{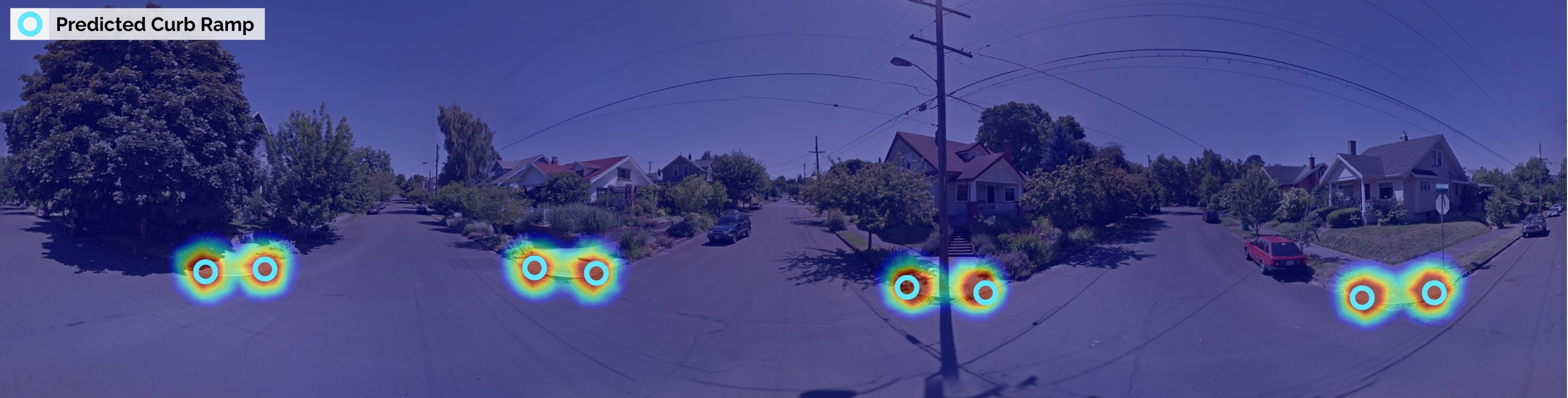}
  \caption{A heatmap generated by our Stage 2 curb ramp detection model. Peaks of the heatmap, which represent predicted curb ramp points, are circled in blue.}
  \label{fig:full_pano}
\end{figure*}

While Stage 1 produces a large-scale dataset of auto-generated curb ramp labels on GSV panos, in Stage 2, we show how this generated dataset can be used to train a state-of-the-art curb ramp detection model using pixels alone--far exceeding prior work~\cite{haraTohmeDetectingCurb2014, weldDeepLearningAutomatically2019}. Crucially, unlike Stage 1, which requires government data, the Stage 2 model works on any city with streetscape imagery available.

\subsection{Model Architecture}

For the computer vision model, we again use the ConvNeXt V2~\cite{wooConvNeXtV2Codesigning2023} architecture, specifically the \textit{base} model variant. To speed up convergence and improve performance, we load weights from a publicly available ConvNeXt V2 model that is pretrained on the ImageNet-1k dataset~\cite{dengImageNetLargescaleHierarchical2009}.

To enable point-based object detection, we modify the architecture's output layers to perform heatmap regression. We replace the classification head with a module consisting of a 3~×~3 convolution, a ReLU activation, and a bilinear upsampling layer, followed by a final 1~×~1 convolution that produces a single-channel heatmap, as shown in \cref{fig:full_pano}. Our model accepts a 4096~×~2048px panorama and outputs a scaled-down 1024~×~512px heatmap of probable curb ramp points. We represent each label in the heatmap by centering a 2D gaussian kernel ($\sigma = 10.0$) at its corresponding location in the image. This heatmap regression approach is commonly used in human pose estimation~\cite{xiaoSimpleBaselinesHuman2018}, but here we use it for point-based object detection. To extract the predicted points, we identify peaks in the heatmap whose maximum value is above a certain threshold (currently set to 0.55). This threshold can be tweaked depending on the user's desired balance of precision and recall.

\subsection{Training}
We use the same 70\% training (150,063 panos), 20\% validation (42,875 panos), and 10\% test split (21,438 panos) described in \cref{sec:stage_one}. During the training process, we augment our data by randomly applying a horizontal flip, as both our own testing and prior work~\cite{shortenSurveyImageData2019} has demonstrated that this increases performance. Our loss function is pixel-wise mean squared error. We use the Adam optimizer~\cite{kingmaAdamMethodStochastic2017} with a learning rate of $1.0 \times 10^{-5}$. We trained the model for a single epoch on 16 NVIDIA L40s GPUs (distributed across four nodes). Our batch size was limited to one, as larger batch sizes were infeasible due to VRAM limitations.

\begin{figure}
  \centering
  \includegraphics[width=3.25in]{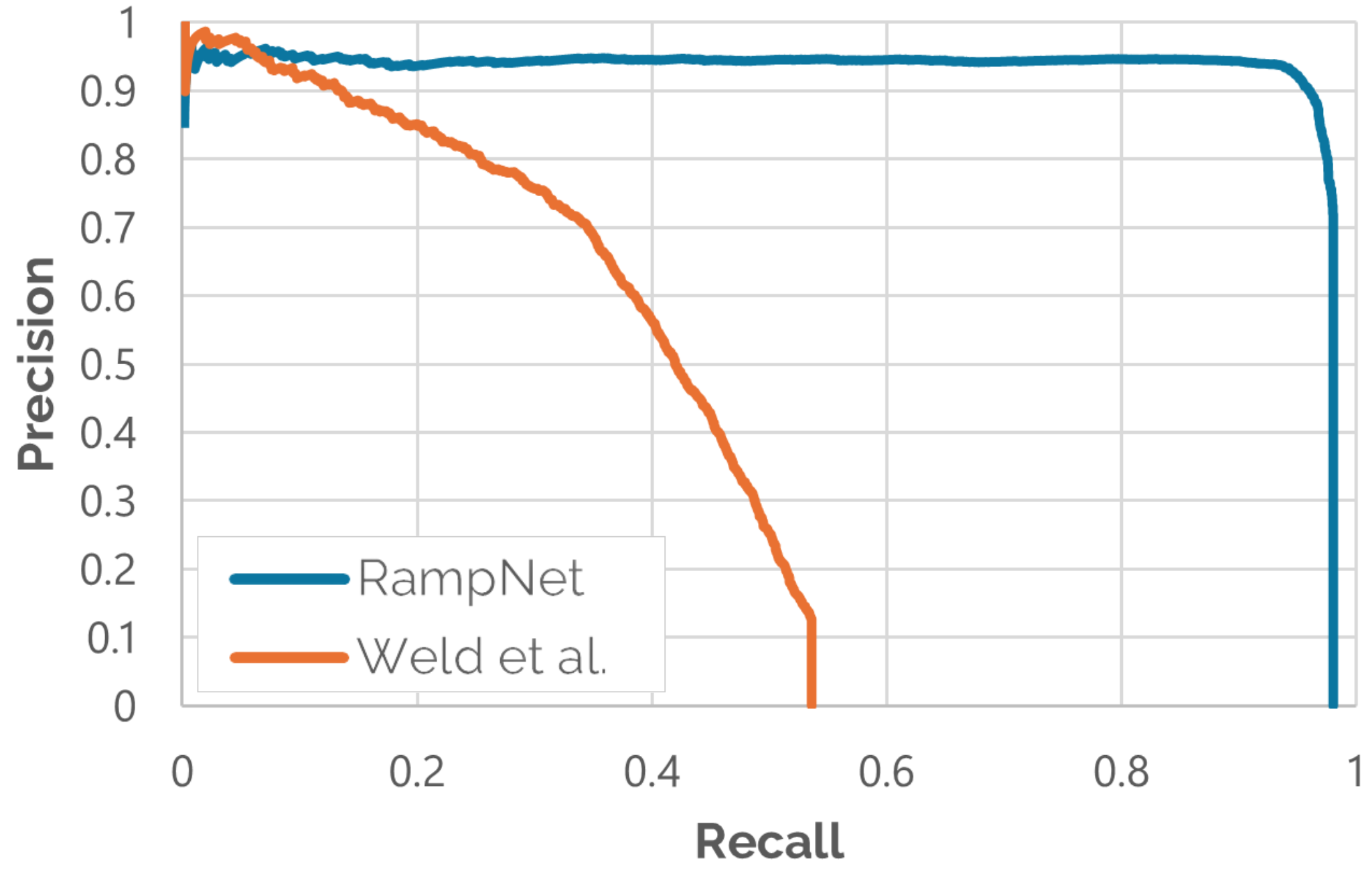}
  \caption{Precision-recall curves of \sysname's curb ramp detection model and the previous state-of-the-art~\cite{weldDeepLearningAutomatically2019}. Both are benchmarked against our manually-labeled dataset.}
  \label{fig:pvr}
\end{figure}

\subsection{Results}
We then tested our trained model against the 1,000-panorama ground truth dataset (with 3,919 manually labeled curb ramps). We report an average precision (AP) of 0.924. To contextualize these results, we compared to the previous state-of-the-art curb ramp detection system released by Weld \etal~\cite{weldDeepLearningAutomatically2019}. Using their released model checkpoints and code~\cite{ProjectSidewalkSidewalkcvassets192025}, we evaluated their curb ramp detection system on our manually labeled dataset, finding that they achieve 0.380 AP, far lower than our 0.924 AP (see \cref{tab:performance_comparison,fig:pvr}). 

Further, we also tested our trained model against the test split of our auto-generated dataset (containing 21,438 panos and 82,055 labels). We report an AP of 0.873. However, we consider this benchmark to be less reliable than our above comparison with the manually-labeled ground truth, due to the higher likelihood of there being errors in the auto-generated dataset.

Naturally, because we are eliminating one input (the government curb ramp data), we might expect lower performance in Stage 2 than in Stage 1. However, we find this performance gap to be surprisingly small, with our Stage 2 model essentially matching the performance of our Stage 1 dataset. When benchmarking against the 1000-panorama ground truth dataset, our Stage 1 generated dataset had a recall of 92.5\% and a precision of 94.0\% versus our Stage 2 model's precision of 93.9\% at the same recall threshold. 

\begin{table}[t]
  \small 
  \centering
  \begin{tabular}{@{}lll@{}}
    \toprule
    & Weld \etal~\cite{weldDeepLearningAutomatically2019} & \sysname Stage 2 \\
    \midrule
    Input Type & Depth+Image+Geo Data & Image Data \\
    Dataset Used & Project Sidewalk~\cite{sahaProjectSidewalkWebbased2019} & Auto-Generated \\
    Avg. Precision & 0.3803 & 0.9236 \\
    \bottomrule
  \end{tabular}
  \caption{Performance comparison of our curb ramp detection model against the previous state-of-the-art. Average precision is calculated by benchmarking against our manually-labeled subset.}
  \label{tab:performance_comparison}
\end{table}

%% file: sec/6_discussion.tex
\section{Discussion}
\label{sec:discussion}

In this paper, we introduced \textit{\sysname}, a custom two-stage pipeline for bootstrapping curb ramp detection in streetscape images using open government metadata and deep learning. Below, we discuss limitations, suggest directions for future work, and describe the impact and potential applications of automatic curb ramp detection.

\subsection{Limitations}

While our dataset includes more than 840,000 curb ramp labels, many of the labels are duplicates of the same physical curb ramp captured from different panorama viewpoints. Indeed, an individual curb ramp appears in 4.5 different panoramas on average. While still useful for training, these repeated views may offer less value than entirely unique curb ramps. During the dataset splitting process, researchers must take care to avoid data leakage by ensuring that the same curb ramp is not distributed across multiple splits, a precaution taken in our work.

As with any deep learning system, our dataset and model contain biases that have the potential to limit generalizability. For example, our dataset was generated from just three U.S. cities. While we strategically selected these locations for their diverse curb ramp styles, the limited coverage area likely introduces regional bias. While we believe our dataset is sufficiently diverse to support the detection of curb ramps in the United States, more work is needed to create datasets for other regions, where curb ramp style differs significantly.

Our system is also limited by its dependence on streetscape imagery, which can sometimes be unavailable or outdated. While up-to-date streetscape imagery is generally available in major U.S. cities, coverage in rural areas, small towns, or rapidly changing neighborhoods can be sparse or outdated, reducing our system's accuracy and timeliness. Indeed, a recent study by Kim and Jang, which analyzed 45 small- and medium-sized cities, found that 44\% of commute routes lacked adequate GSV imagery and that there was significant temporal variability in the coverage~\cite{kimExaminationSpatialCoverage2023}.

\subsection{Future Work}

Our long-term goal is to enable fully automatic urban accessibility assessment. We have addressed a key component of that goal by developing a method for automatic curb ramp detection. However, other unsolved problems remain.

To assess curb ramp accessibility, cities must (1) know \textit{where} curb ramps are and (2) be able to evaluate key quality factors (\eg, presence of tactile warnings, steepness). While our work addresses the first requirement by enabling the automatic detection of curb ramps, further work is needed to enable automatic quality assessment. We posit that our auto-translation technique could be extended for this task, as quality information is often included in government-provided curb ramp metadata. Although \sysname, like Project Sidewalk \cite{sahaProjectSidewalkWebbased2019}, outputs single-point labels, future work should explore generating bounding boxes to provide richer spatial information crucial for quality assessment (\eg, ramp width and alignment).

While the focus of our work is auto-translating location coordinates to image pixel coordinates, the reverse process is equally important. Current city workflows are reliant on location data rather than the image pixel coordinates that our model outputs. The introduction of a system to translate back to location coordinates would greatly enhance the usability of our curb ramp detection model. 

Lastly, curb ramps are not the only urban accessibility feature that could benefit from automatic detection and analysis. Other features (\eg, pedestrian signals, missing curb ramps, path obstacles, surface problems) play a similarly vital role in ensuring urban accessibility. We encourage future work to explore ways of automatically detecting and assessing these additional features to provide a more comprehensive understanding of accessibility conditions.

\subsection{Impact on Urban Accessibility}

Our research aids in urban accessibility planning by enabling a low-cost, scalable method for curb ramp detection. Our dataset achieves 92.5\% recall and 94.0\% precision, and our curb ramp detection model reaches 0.924 AP, significantly outperforming prior work and, for the first time, achieving near-human-level quality.

We envision a future where whole cities can be audited in a single day instead of over the course of multiple months. By reducing time and cost requirements, we allow more cities to conduct crucial accessibility audits. We also give accessibility advocates a powerful tool to hold cities accountable to the legal requirements defined by the \textit{Americans with Disabilities Act}~\cite{AmericansDisabilitiesAct}.

%% file: sec/7_conclusion.tex
\section{Conclusion}
\label{sec:conclusion}

In conclusion, \sysname advances research in automatic curb ramp detection by (1) offering a novel technique for auto-translating open government datasets into GSV pano labels, by (2) producing the largest and most comprehensive curb ramp detection dataset, and by (3) establishing key performance benchmarks for automatic curb ramp detection (0.924 AP). We believe these contributions provide a foundation for fully automatic urban accessibility assessment and future research in the area.


%% file: sec/8_acknowledgements.tex
\section{Acknowledgments}
\label{sec:acknowledgments}

This work was supported by NSF SCC-IRG \#2125087 and NSF OAC \#2411222.